\documentclass[a4paper,conference]{IEEEtran}
\usepackage{amsmath,amssymb,amsfonts}
\usepackage{algorithmic}
\usepackage{threeparttable}
\usepackage{subcaption}
\usepackage{graphicx}
\usepackage{marvosym}

\ifCLASSINFOpdf
\else
\fi
\begin{document}

%
\title{Pay attention to emoji: Feature Fusion Network with EmoGraph2vec Model
for Sentiment Analysis}

\author{
  \IEEEauthorblockN{Xiaowei Yuan\textsuperscript{1,2}, Jingyuan Hu\textsuperscript{1}, Xiaodan Zhang\textsuperscript{1}\textsuperscript{(\Letter)}, Honglei Lv\textsuperscript{1}\\
  \IEEEauthorblockA{
    \textit{\textsuperscript{1}Institute of Information Engineering, Chinese Academy of Sciences, Beijing, 100093, China} \\
    \textit{\textsuperscript{2}School of Cyber Security, University of Chinese Academy of Sciences, Beijing, 100049, China}\\
    E-mail: \{yuanxiaowei, hujingyuan, zhangxiaodan, lvhonglei \}@iie.ac.cn 
 }}}


%


\maketitle

\begin{abstract}
With the explosive growth of social media, opinionated postings with emojis have increased explosively.
Many emojis are used to express emotions, attitudes, and opinions.
Emoji representation learning can be helpful to improve the performance of emoji-related natural language processing tasks, especially in text sentiment analysis.
However, most studies have only utilized the fixed descriptions provided by the Unicode Consortium without consideration of actual usage scenarios. As for the sentiment analysis task, many researchers ignore the emotional impact of the interaction between text and emojis. It results that the emotional semantics of emojis cannot be fully explored.
In this work, we propose a method called EmoGraph2vec to learn emoji representations by constructing a co-occurrence graph network from social data and enriching the semantic information based on an external knowledge base EmojiNet to embed emoji nodes. Based on EmoGraph2vec model, we design a novel neural network to incorporate text and emoji information into sentiment analysis, which uses a hybrid-attention module combined with TextCNN-based classifier to improve performance.
Experimental results show that the proposed model can outperform several baselines for sentiment analysis on benchmark datasets. Additionally, we conduct a series of ablation and comparison experiments to investigate the effectiveness and interpretability of our model.
\end{abstract}


%
\IEEEpeerreviewmaketitle

\section{Introduction}
With the rapid growth of usage of emoji in all social media posts and messages, it has become a part of culture in computer-mediated communications \cite{HuGSNL17}. In fact, the Oxford Dictionary named ‘face with tears of joy' as the word of the year in 2015\footnote{https://goo.gl/6oRkVg}. For most natural language processing (NLP) tasks, the fundamental step is text representation, by converting unstructured text into structured information, so that it can be calculated in many tasks such as text classification and sentiment analysis. Word embedding is the mainstream method to represent words in finite-dimensional vector space by word2vec \cite{Mikolov} or GloVe \cite{Glove}. Similarly, emoji representation learning can be helpful to improve the performance of emoji-related NLP tasks especially in text sentiment analysis \cite{Wijeratne}. The ability to automatically process and interpret text combined with emoji will be vitally important as the ubiquity of emoji. Representing the emoji meanings like word embedding models can be used to encode emoji into vectors, which we call emoji embedding models.


Previous study \cite{HuGSNL17, Cramer} shows that the usage of emojis has sentiment effects on plain text, which can even dominate the overall emotional polarity. 
For example, the sentiment valence of text \textit{“Today is a rainy day."} is originally neutral. If an emoji 'grin' (or 'sob') was added in the end, however, the sentiment would be totally changed.
And we also notice that some emojis are ambiguous in their meanings depending on different contexts  and the co-occurrence emojis appearing in a sentence have the likeness of their sentiment, interpretation or intended use. It is of great importance to understand the emotional semantics of emoji and analyze the overall sentiment of the sentence. 
Consider two tweets, \textit{1) The white curtains blew in the wind like 'ghost', it's a little scary 'scream'. 2) I love 'ghost' movie 'heart-eyes' 'heart-eyes'.} The intended emotional meaning of 'ghost', with different emojis followed, varies from negative to positive. This also reflects that the emotional information in emojis is not immutable.
Therefore, we propose a model called EmoGraph2vec to obtain better performance on emoji representation learning. We utilize the co-occurrence of emoji in “big data” as an underlying feature to construct non-euclidean structured data to learn emoji embeddings in a graph with a variational graph autoencoder (VGAE) \cite{VGAE}. The semantics of emojis are provided by a sense inventory called EmojiNet \cite{EmojiNet} to add node attributes and edge weights.

With the explosive growth of social media, opinionated postings with emojis have increased explosively. Social media posts contain a vast amount of valuable emotional information and have become a hot research target for sentiment analysis \cite{Cambria}.
Text with emojis could resolve ambiguity in written communication and allow people to vividly convey their feelings in a concise way. Therefore, fusing text with attached emojis can be more conducive to sentiment analysis task, which also could be viewed as a bi-modal (text + emoji) information processing. The most common method to analyze the text containing emojis is to utilize emojis as an important feature \cite{10.1145, Lou}. Nevertheless, they are not fine-grained enough in feature fusion, which means they failed to explore fully the emotional impact of emojis on the text. In this paper, we propose a hybrid-attention network to learn the mutual emotional features between text and emojis. Then we send fused features into a TextCNN\cite{Kim}-based classifier to predict the sentiment labels.


In summary, the main contributions of our work are as follows:
\begin{itemize}
\item \small We propose an emoji representation learning method named EmoGraph2vec to learn emoji embeddings in a graph network. We utilize a large-scale unlabeled corpus to form an emoji co-occurrence graph network and adopt a VGAE to make use of network structure and node attribute information to obtain emoji embeddings based on EmojiNet.
\item \small We propose a novel neural network framework to incorporate text and emoji information into sentiment analysis, which uses a hybrid-attention network to fuse the bi-modal features and combines with a TextCNN-based classifier to predict sentiment labels.
\item \small We confirm the effectiveness of our model for sentiment analysis on several emoji-rich datasets. We also visualize the Emograph2vec compared with other method to confirm the interpretability of our emoji representation method.
\end{itemize}

\section{Relate Work}
\subsection{Sentiment Analysis}
As a significant branch of natural language processing (NLP), text sentiment analysis aims to mine and analyze the emotions, opinions, and attitudes from texts. 
The rapid development of deep learning has played an important role in boosting the development of sentiment analysis researches. Socher et al. \cite{Socher} applied Recursive Neural Network to text sentiment classification with the consideration of the syntactic structure information; Santos et al. \cite{dos-santos-gatti-2014-deep} proposed Character to Sentence Convolutional Neural Network to analyze sentiment on the short text.

\subsection{Emoji in Sentiment Analysis}
Many studies use emojis as heuristic information in social texts \cite{Davidov,Go,Li}, where emojis serve for unsupervised learning in a large number of unlabeled data. 
Emojis can also be regarded as important semantic features about emotions towards the recipient or subject \cite{kaw} for sentiment classification. For instance, Tian et al. \cite{ZTian} used a Bi-directional Gate Recurrent Unit Attention network, which integrated the emotional polarity of emoji and embedded it as a feature into the model. But they failed to reflect the emotional impact of emojis on the text. Lou et al. \cite{Lou} constructed an emoji-based bidirectional long short-term memory (Bi-LSTM) model, which combined the attention mechanism to weigh the contribution of each word on the emotional polarity based on emoji. 
But their model only analyzes the microblog data that contains a single emoji and cannot be generalized to process multiple types of emojis. Yuan et al. \cite{Yuan} proposed an emoji-based co-attention network to deal with the text and attached multiple emojis. But all the works only learned the emoji features through fixed emoji names without consideration of actual usage scenario.
\subsection{Emoji Representation Learning}
Barbieri et al. \cite{Barbieri} trained an emoji embedding model on tweets to seek an understanding of emoji meanings from how emoji are used in a large collection of tweets. Eisner et al. \cite{Eisner} used a pre-trained word embedding model, and applied it to emoji names to learn an emoji embedding model which they called emoji2vec. 
Wijeratne et al. \cite{Wijeratne} learned the distributional semantics of the words in emoji definitions to model the emoji meanings extracted from EmojiNet. But they failed to explore the emoji meanings in the actual usage scenarios. 
Illendula et al. \cite{Illendula} utilized a large Twitter corpus which has emojis in them and built an emoji co-occurrence network, and trained a network embedding model to embed emojis into a low dimensional vector space. 
But they only utilized the emoji frequency and ignored their textual meanings as node information which can enhance the emoji semantic features.

\section{Model}
\begin{figure*}[htbp]
  \centering
  \includegraphics[width=.8\linewidth]{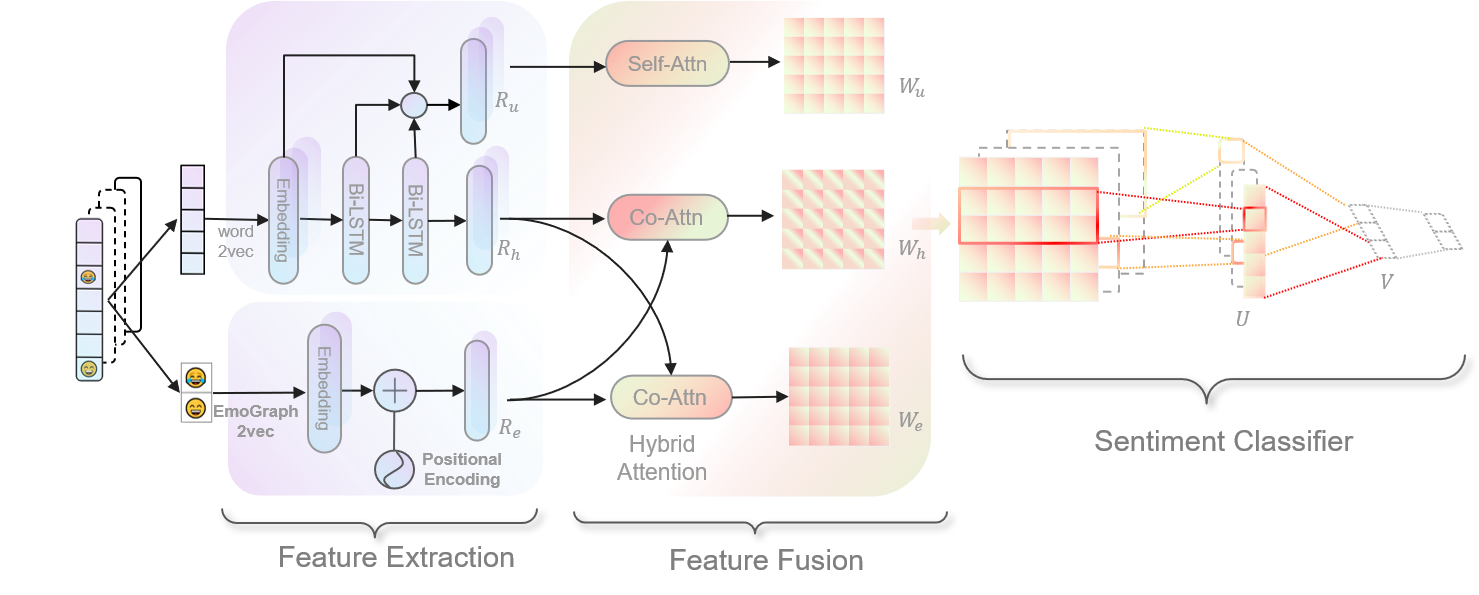}
  \caption{The architecture of the whole model}
  \label{fig:model}
\end{figure*}

For sentiment analysis, our model consists of several major procedures: feature extraction, feature fusion and sentiment classifier (Fig. \ref{fig:model}).

We will illustrate the details in the following section.

\subsection{Emoji Feature Extraction}
Given posts with text and emojis, we first use our emoji embeddings model EmoGraph2vec to encode emoji into vectors, so as word2vec does. Then we equip the embeddings with additional information about the emoji position in the sentence since our model treats emoji as a separate input, which derives from the original sentence, for the model to make use of the order of the words and emojis.

\subsubsection{EmoGraph2vec}
\begin{figure}[hb]
  \centering
  \includegraphics[width=.9\linewidth]{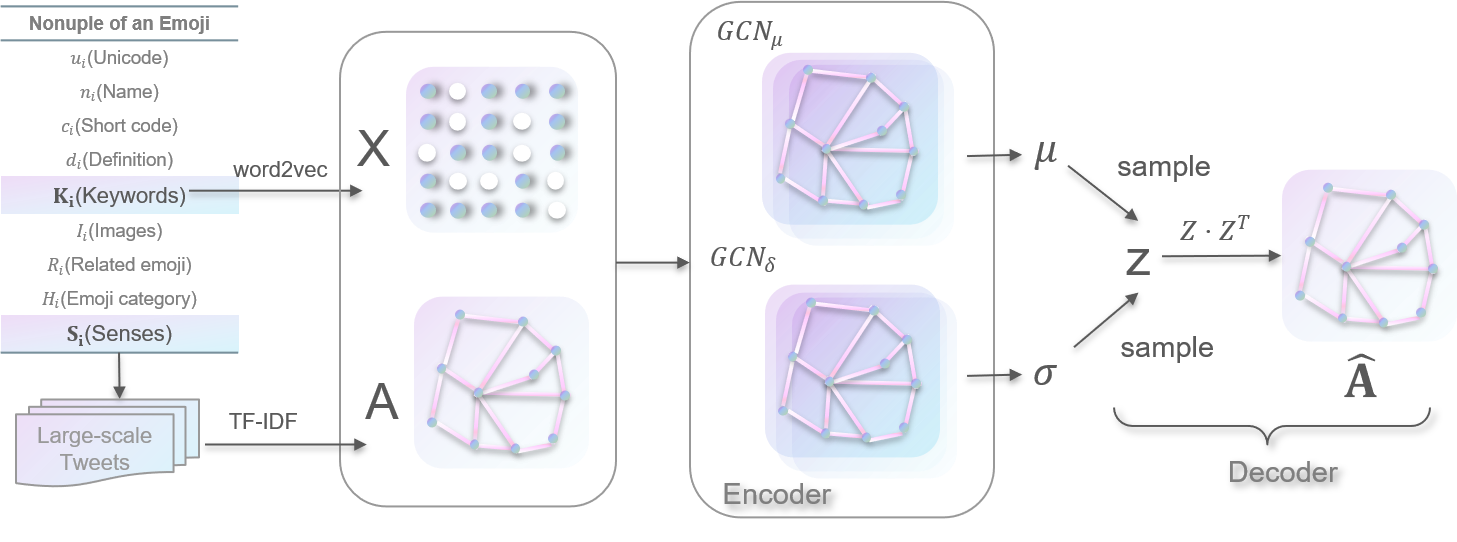}
  \caption{The architecture of EmoGraph2vec model}
  \label{fig:emograph2vec}
\end{figure}

Based on large-scale Twitter data (\ref{unlabeled_data}), we extract the emoji co-occurrences from it to learn emoji features with the EmojiNet resource. Different from traditional methods that treat the emoji co-occurrences as sequence data, we utilize the co-occurrence information from the whole corpus as non-Euclidean data to construct an undirected network graph. 
\paragraph{EmojiNet} EmojiNet is a machine readable sense inventory for emojis created by Wijeratne et al \cite{EmojiNet}. It consists of 12,904 sense labels over 2,389 emojis, where each emoji $e$ is represented as a nonuple $(u,n,c,d,K,I,R,H,S)$. The meaning of each element is shown in the Figure \ref{fig:emograph2vec}.
We utilize the EmojiNet to transform the Unicode emoji representation into textual meanings.
Specifically, $K$ and $S$ elements of each emoji nonuple are used to calculate edge weight and the node attribution in the graph network. 

\subsubsection{Graph Initialization}
Each tweet forms a polygon of $n$ nodes where $n$ represents the number of unique emojis. Let us denote the graph as $G=(V,E)$, where $v\in V$ denotes one kind of emoji, and $N=|V|$ equals the number of all emojis. The nodes are connected when they appear in the same tweet. 

The edges of the network are weighted by the co-occurrence frequency and similarity of two emoji nodes. 
We extract senses $S$ of each emoji $e_i$ from EmojiNet as node text to calculate the TF-IDF (Term Frequency–Inverse Document Frequency) vectors $v_i$. Then, similarity scores between node pairs $(e_i,e_j)$ are computed by cosine similarity. We hypothesize that the emoji pair that has high co-occurrence frequency may have a strong emotional connection and the similarity can quantify the emotional semantic distance. Based on co-occurrence frequency and similarity, the product of these two indicates edge weight in weighted adjacent matrix $A$, which can determine the mutual influence of emotional polarity between emoji pairs. 
\begin{equation}
    A[i,j]=cosine(v_i,v_j)*frequency(e_i,e_j)
\end{equation}

To enrich the semantic information in the network, we also utilize the $K$ element of emoji $e$ into this graph as node attributes. 
We embed the keywords of each emoji by word2vec. The word vectors $k_i$ of all words in the emoji senses are averaged to form the final attribute $x_e$.
\paragraph{Graph Embedding}
To incorporate the network structure and node attribute information, we use an unsupervised learning method VGAE to learn node embeddings. This model uses a two-layer graph convolutional network (GCN) encoder and an inner product decoder. The encoder produces the distribution of vectors, including mean $\mu$ and variance $\sigma$, from which stochastic latent variable $z_i$ is obtained by sampling.
\begin{gather}
q(z_i|X,A)=\mathcal{N}(z_i|\mu_i,diag({\sigma}^2_i))
\end{gather}
Here, $A$ is a weighted adjacency matrix of $\mathcal{G}$, $X$ represents the node attribute. And the decoder is given by an inner product between latent variables: 
\begin{gather}
    p(A|Z)=\begin{matrix} \prod_{i=1}^N \begin{matrix} \prod_{j=1}^N p(A_{ij}|z_i,z_j) \end{matrix} \end{matrix} 
\end{gather}
where $Z$ is the emoji embedding matrix we needed.
\subsubsection{Positional Encoding}
Positional encoding are useful for the emojis because they give the emoji features a sense of where the emoji would have been. We combine position encoding with the emoji embeddings to form the input emoji features $R_e=(e_1+pe_1,e_2+pe_2,...,e_n+pe_n)$. The dimension of positional encoding is the same as the emoji embeddings so as to sum the both.

In our model, we utilize sine and cosine functions for different positions\cite{Jones}:
\begin{gather}
   PE_{(pos,2i)}=sin(\frac{pos}{1000^{2i/d_{model}}}) \\
   PE_{(pos,2i+1)}=cos(\frac{pos}{1000^{2i/d_{model}}})
\end{gather}
Where $pos$ is the absolute position of emoji in original sequence, $i$ is the dimension. Given the mathematical properties of sinusoidal and cosine functions, these two functions can also learn to attend by relative positions, since $PE_{pos+t}$ can be represented as a linear transformation of $PE_{pos}$.
                                                        
\subsection{Text Feature Extractor}
We obtain the word embeddings by the pre-trained word vectors over the Google News dataset\footnote{https://goo.gl/QaxjVC}.  Then the plain text can be represented as $X=[x_1,x_2,...,x_L]$. Since LSTM can overcome the problem of gradient vanishing and explosion with the capability to learn long-range dependencies in sequences, our text feature extractor adopts two stacked Bi-LSTM layers to learn the text representation bidirectionally.
And we concatenate the hidden vectors from both directions to represent every single word as the output $h_l$ of the layer.
The second bi-directional LSTM layer takes the output of the previous one as its input $H_{1}=[h_{11},h_{21},..,h_{L1}]$, and computes unit stats of network in the same pattern
before producing the output $H_{2}=[h_{12},h_{22},..,h_{L2}]$ as $R_h$.
Through a skip-connection, the outputs of the below three layers (the embedding layer and the two bi-directional LSTM layers) are concatenated as $R_{u}=[X,H_{1},H_{2}]$.
\subsection{Feature Fusion}
For a post containing text and emoji, not all words contribute equally to express sentiments, we need to lead the model to attend to sentiment-guided words. Furthermore, different emojis are also related to different emotional semantics, which depends on the context. Also text has a strong emotional impact on the sentiment conveyed by the emoji. To this end, we develop a novel hybrid-attention module to model the mutual influence between text and emoji based on self-attention unit and its variant co-attention unit.

\begin{figure}[ht]
    \centering
    \begin{minipage}[t]{0.45\linewidth}
        \includegraphics[width=\textwidth]{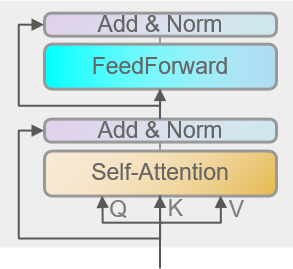}
        \caption{self-attention unit}
        \label{fig:self}
    \end{minipage}%
    \hfill%
    \begin{minipage}[t]{0.45\linewidth}
        \includegraphics[width=\textwidth]{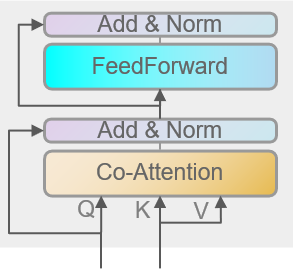}
        \caption{co-attention unit}
        \label{fig:co}
    \end{minipage}
\end{figure}
Self-attention unit is a crucial part in transformer \cite{Transformer} as showed in Figure \ref{fig:self}, which consists of self-attention function and a fully connected feed-forward network with a residual connection and normalization. The input vectors are used to compute three matrices $Q$, $K$, and $V$ (queries, keys, and values).
And the attention distributions over $V$ are determined by the dot-product similarity between $Q$ and $K$:
\begin{equation}
    Attention(Q,K,V)=softmax(\frac{QK^T}{\sqrt{d}})V
\end{equation}
where $d$ is the dimension of of the output feature.
The resulting weight-averaged value vector forms the output of the attention block. The fully connected feed-forward network consists of two linear transformations with a ReLU activation function in between.
\begin{equation}
    FFN(x)=max(0,xW_1+b_1)W_2+b_2
\end{equation}

The co-attention unit is a variant from self-attention \cite{Jiasen}.The only distinction is that the queries are from emoji (or text) features while keys and values are from text (or emoji) features. This unit produces an attention-pooled feature for one modality conditioned on another modality. If $Q$ is computed by text and $K$ and $V$ computed by emoji, the unit measures text-guided attention to decide which emoji should weigh more. If $Q$ comes from emoji and $K$ and $V$ come from text, this unit measures emoji-guided attention to weigh each word. The whole process we design is to simulate the mutual influence between text and emoji.

\subsection{Sentiment Classifier}
After the hybrid attention module, we obtain the bi-modal feature matrixes $W_u\in\mathbb{R}^{d*L}$, $W_e\in\mathbb{R}^{d*N}$ and $W_h\in\mathbb{R}^{d*L}$ ($d$ is the embedding dimension, $L,N$ is the length of text and emojis) fused features of text and emoji. The three matrixes are treated as three single-channel input into a TextCNN classifier to predict the probability distribution of sentiment labels. We use three sizes $(k_1,k_2,k_3)$ of kernel sets $S_i=[s_{i1},s_{i2},…,s_{in}]$, $s_i\in\mathbb{R}^{k_i*d} (i=1,2,3)$, that respectively map the input $W_u$, $W_e$ and $W_h$ to $n$ new feature maps $U_i=S_i*W_i, U_i\in\mathbb{R}^{k'*1}$. Here, $*$ denotes convolution. When $i$ takes different values, $W_i$ represents $W_u, W_e$ or $W_h$.
We then apply a max-over-time pooling operation to take the maximum value for each map, and form the new representation $V_u, V_e, V_h\in\mathbb{R}^n$. The three outputs are concatenated to be passed to a fully connected softmax layer whose output is the probability distribution over sentiment labels.

\section{Experiment}
\subsection{Dataset}
\subsubsection{Labeled Dataset}
The labeled data are collected from multi-source social media platforms and cover multiple domains to minimize biases. More details about the datasets (MSD, TD, and ERD) are shown in Appendix A \footnote{In supplementary material}. We use the part of the data that contains the emojis, of which the amount is limited. 

\subsubsection{Unlabeled Dataset}\label{unlabeled_data}
To train the EmoGraph2vec model, we use a large-scale unlabeled data of Tweets named EmojifyData\footnote{https://www.kaggle.com/rexhaif/emojifydata-en}. This dataset contains 18 million English tweets, all with at least one emoji included. Based on it, we can learn emoji representations in EmoGraph2vec to obtain emotional semantic information in their embeddings. 


\subsection{Implementation Details}
Our model is trained using the PyTorch library \cite{PyTorch} on a cuda GPU. Appendix B provides a detailed description of our experimental configuration. In VGAE, the number of units in hidden layer 1 is set to 256, and layer 2 is 300. The training procedure runs for 50 epochs to learn emoji representations on EmojifyData. 
We train our model in the sentiment analysis task with 20 epochs. 

\subsection{Baselines and Performance Comparison}
To evaluate the performance of our model, we employ several representative baseline methods: TextCNN \cite{Kim}, Att-BiLSTM \cite{Zhou}, EA-Bi-LSTM \cite{Lou} and ECN\cite{Yuan}. EA-Bi-LSTM (based on \underline{E}moji-\underline{A}ttention and \underline{BiLSTM}) and ECN (\underline{E}moji-based \underline{C}o-attention \underline{N}etwork) are the latest work that designed for the text containing emojis. 
For each method, we use three methods to embed emojis. The line without symbol uses the EmoGraph2vec method, others use emoji2vec \cite{Eisner} and the Unicode characters given by the Unicode Consortium.
The emoji2vec embeddings are publicly released\footnote{https://github.com/uclmr/emoji2vec}. 

\begin{table}[t]
\footnotesize
\begin{center}
\begin{threeparttable}
\caption{The accuracy of our model and baseline methods on different dataset}
\setlength{\tabcolsep}{4mm}
\begin{tabular}{c|c|c|c}
\hline
\textbf{Models} & \textbf{MSD}& \textbf{TD}& \textbf{ERD}\\
\hline
TextCNN& 0.8329& 0.7273& 0.6856 \\
TextCNN\dag& 0.8258& 0.7202& 0.6719\\
\textbf{TextCNN*}& \textbf{0.8524}& \textbf{0.7378}& \textbf{0.7046} \\
\hline
Att-BiLSTM& 0.8217& 0.7098& 0.6666\\
Att-BiLSTM\dag& 0.8243& 0.7198& 0.6700 \\
\textbf{Att-BLSTM*}& \textbf{0.8301}& \textbf{0.7203}& \textbf{0.6792} \\
\hline
\hline
\textbf{EA-Bi-LSTM}& \textbf{0.8607}& \textbf{0.7483}& \textbf{0.7278}\\
EA-Bi-LSTM\dag& 0.8527& 0.7470& 0.7025 \\
EA-Bi-LSTM*& 0.8301& 0.7413& 0.6835 \\
\hline
\textbf{ECN}& \textbf{0.8672}& \textbf{0.7579}& \textbf{0.7335} \\
ECN\dag& 0.8552& 0.7464& 0.7256 \\
ECN*& 0.8213& 0.7295& 0.6929\\
\hline
\textbf{Our model}& \textbf{0.8815}& \textbf{0.7703}& \textbf{0.7627}\\
Our model\dag& 0.8567& 0.7595& 0.7464\\
Our model*& 0.8335& 0.7442& 0.7401\\
\hline
\end{tabular}
\label{tab1}
\begin{tablenotes}
    \item[\dag] use the emoji2vec method to obtain emoji representations
    \item[*] use the Unicode characters as emoji representations
\end{tablenotes}
\end{threeparttable}
\end{center}
\end{table}

In Table \ref{tab1}, we can see our model outperforms all other baseline methods with the EmoGraph2vec model. 
The results prove that our proposed method is more effective than the traditional deep learning methods (TextCNN, Att-BiLSTM), which do not pay attention to emojis. 
Looking more closely, EA-Bi-LSTM and ECN achieve a certain improvement than the former methods (regardless of EmoGraph2vec or emoji2vec). It demonstrates the significance of incorporating emojis and text to analyze sentiment. 
And it is remarkable to find that our model obtains higher accuracy than other baselines. What's more, EmoGraph2vec can achieve higher accuracy compared with emoji2vec and unicode in the latter three models that focus on emojis.
Especially, when our model is applied on the ERD (Emotion Recognition Dataset), there is a noticeable drop in the overall accuracy. Different from sentiment classification, emotion is more fine-grained and multi-classified, which puts forward higher requirements for the accuracy of the model; secondly, the amount of training set in the ERD dataset is also less than the first two sentiment analysis datasets, which may also have some impact on accuracy.

When using the Unicode characters as emoji representations, one interesting finding is that the accuracy of our model, EA-Bi-LSTM and ECN have decline while TextCNN and Att-BiLSTM have increase. 
A possible explanation for this might be that the former separate text and emojis as two parts of inputs, where pre-trained emoji embeddings can work better. When treating the whole sentence as an integrated input (in TextCNN and Att-BiLSTM), the Unicode characters can be trained with word vectors together to achieve better performance.

\begin{table}[b]
\footnotesize
\begin{center}
\caption{The accuracy of modified models on different datasets}
\setlength{\tabcolsep}{4mm}
\begin{tabular}{c|c|c|c}
\hline
\textbf{Models} & \textbf{MSD}& \textbf{TD}& \textbf{ERD} \\
\hline
N-model& 0.7814&	0.6783&	  0.6709\\

T-model& 0.8356&	0.7215&	 0.7148\\

E-model& 0.8626&	0.7578&	0.7484\\
\hline
\hline
RA1& 0.8512&	0.7448&	 0.7436\\

RA2& 0.8157&	0.7103&	 0.7178\\

RA3& 0.8652&	0.7675&	0.7568\\

\hline
\textbf{Our model}& \textbf{0.8815}& \textbf{0.7703}& \textbf{0.7627} \\
\hline
\end{tabular}
\label{tab:NTE}
\end{center}
\end{table}

\begin{figure*}[t]
    \centering
    \begin{subfigure}{0.4\textwidth}
        \includegraphics[width=\textwidth]{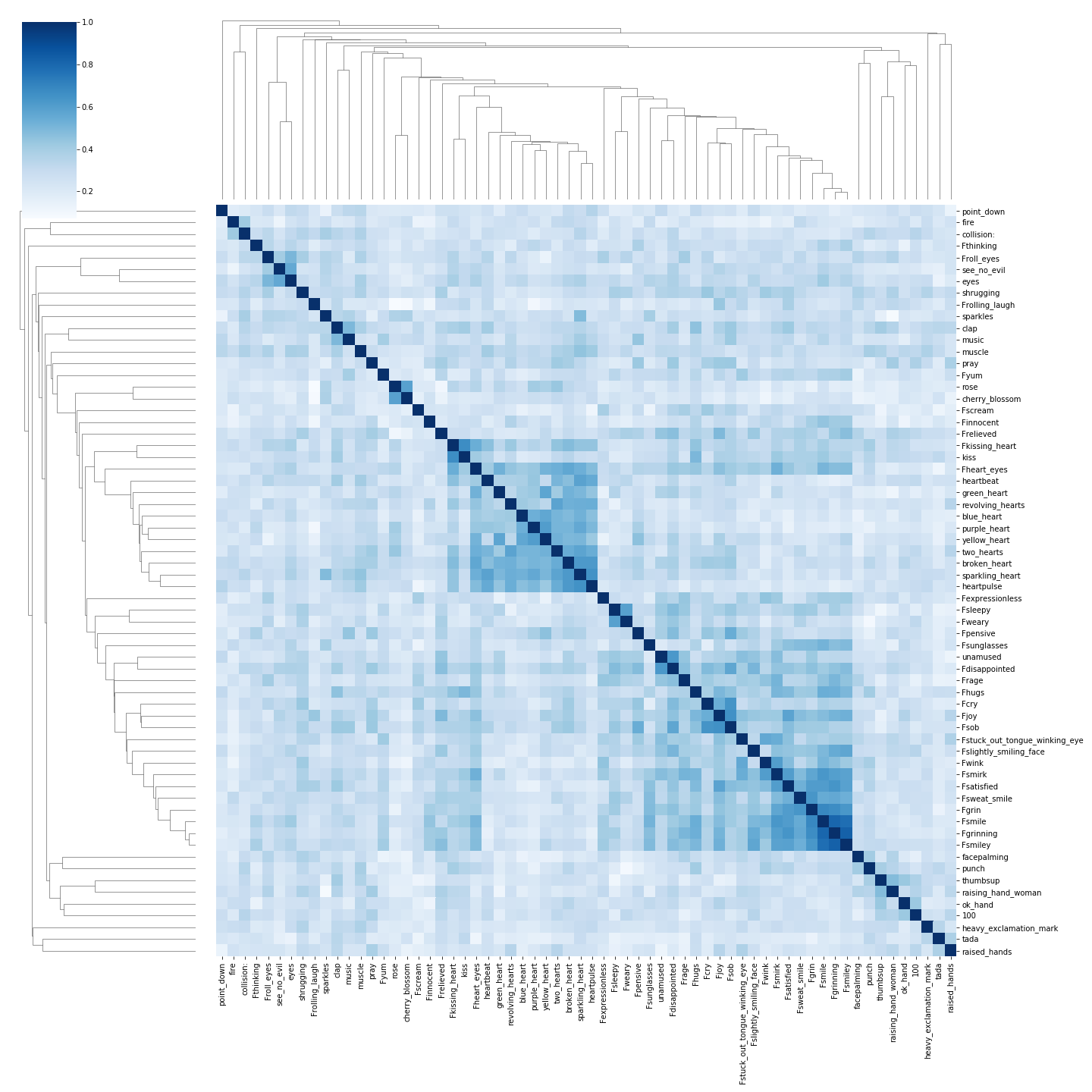}
        \caption{Emoji2vec}
        \label{fig:emo}
    \end{subfigure}
    \hspace{.3in}
    \begin{subfigure}{0.4\textwidth}
        \includegraphics[width=\textwidth]{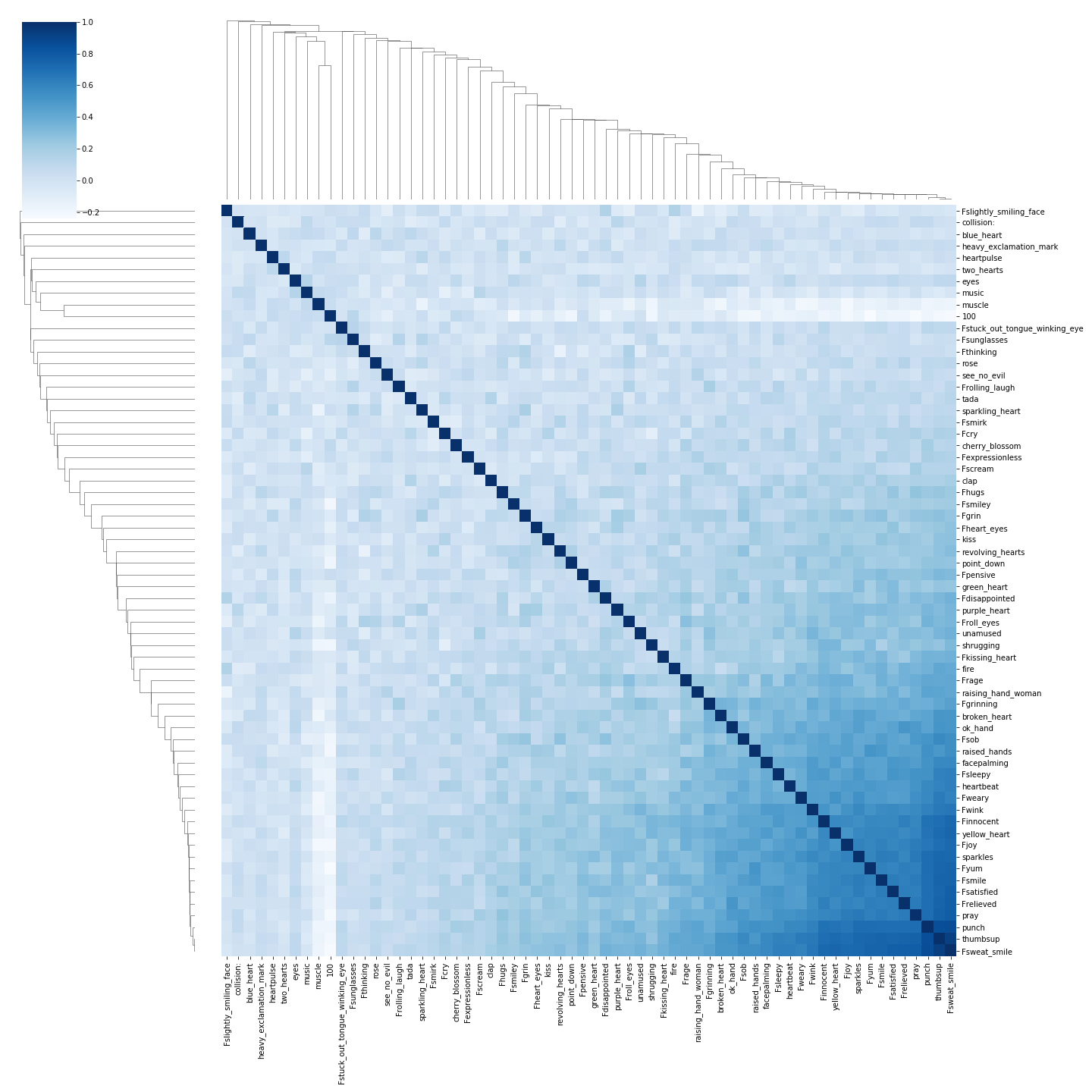}
        \caption{EmoGraph2vec}
        \label{fig:emograph}
    \end{subfigure}
    \caption{Comparison of emoji representations between emoji2vec and emograph2vec. The horizontal and vertical coordinates are emoji names, where 'F' indicates the facial emoji.}
    \label{fig:visual_smap}
\end{figure*}
\subsection{Model Analysis}
\subsubsection{The Power of Emojis}
To further explore the influence of emojis in our model, we conduct the subsequent experiments by removing the inputs of emojis or simplified architecture of the model to evaluate the effectiveness of emojis. N-model detaches the emoji inputs.
T-model removes the text-guided co-attention module of emoji representation learning. E-model removes the emoji-guided co-attention module of text representation learning. 

As shown in Table \ref{tab:NTE}, we find that the complete model significantly outperforms the N-model and T-model on all the datasets, both of which only consider the text features before classification. That demonstrates the plain text does not contain rich emotional semantic information as emojis do occasionally in sentiment analysis. The T-model also outperforms the N-model to a certain degree. This shows emoji-guided text representation learning can effectively improve the ability of the model to learn the emotional semantic. It also explains why our emoji-aware method can achieve better accuracy compared to other baseline methods. The accuracy of the E-model is also higher than T-model and slightly lower than the complete model. Because E-model extracts sentiment information from text-guided emoji representation but fails to capture sentiment patterns of emoji-guided text representation.

\subsubsection{Effectiveness of Co-Attention} 
To further explore the effect of the co-attention mechanism in our proposed method, we compare the complete model with several attention-modified models as follows:
In RA1, we remove the self-attention module and concatenate the text features of each word $R_u$ as text feature matrix $W_u$. In RA2, we replace the text-guided co-attention module with the concatenation of the emoji vectors $E=[e_1,e_2,...,e_N]$ as the emoji representation $W_e$. In RA3, we replace the emoji-guided co-attention module with the concatenation of the text representation $H_{2}=[h_{12},h_{22},..,h_{L2}]$ as the text representation $W_h$. 

As shown in Table \ref{tab:NTE}, it can be seen that the accuracy of the RA2 has dropped sharply in most datasets. RA2 directly changes the attention mechanism of the emoji feature matrix $W_e$ to the concatenation of all emoji vectors, indicating that emojis take most of the weight for the emotional semantic analysis of the model. That means when the model cannot distinguish which emoji dominates the text emotion, the accuracy drops significantly. 
The slight difference of accuracy between the RA1 and RA3 model reveals that when the self-attention of the text is removed, the simplified $W_u$ matrix will further affect the representation of $W_e$ as the emoji feature, and it results in the lower accuracy. While the RA3 retains the first two representations of $W_u$ and $W_e$ matrices, only the last step of the text vector representation is replaced, it has the least impact on the model performance, indicating that the model can still make correct predictions from the $W_e$ matrix with a greater probability.


\subsubsection{Comparison about Emoji Representation Learning}
As shown in the Table \ref{tab1}, when using different emoji embedding methods in sentiment analysis tasks, our model combined with Emograph2vec outperforms other methods in emoji2vec.
To further explore the difference between the two methods, we perform a hierarchical clustering \cite{Ziv} on Top64\footnote{https://home.unicode.org/emoji/emoji-frequency/} most used emojis and visualize the clustering results in Fig \ref{fig:visual_smap}. Each emoji can be meaningfully represented by a low-dimensional vector in the embedding space where similarity can be measured. In the report, the Top64 emojis include facial expressions, gestures, and objects. We suppose that not only facial expression emojis can carry emotional semantic information but also other categories in various scenarios. Since we learn emoji features from social network data, we can learn the sentiment tendency contained in each emoji in a specific scene. We expect that a well-performed representation can embed emojis with similar emotional semantics closely in the vector space no matter it is a facial expression or object. 

The color scale of each cell indicates the similarity between the two emojis. The darker the cell is, the more similar the representations of the two emojis are. As shown in \ref{fig:emo}, the emoji2vec method directly works on Unicode descriptions, which reflects on the clustering result. The facial expression emojis have a high degree of similarity also the similar-shaped (e.g. heart-shaped) emojis do. And the distance between the groups is large, which illustrates that the method strictly classifies emojis literally. 
In the contrast, we can see in the \ref{fig:emograph}, there is no clear boundary to distinguish each kind of emojis. The bottom right corner contains the darkest cells that means these emojis have similar emotional semantics. We can see this area includes expressions as well as gestures and objects (e.g. "sparkle" and "Face\_joy"). 

\section{Conclusions}
Emojis have an extremely high occurrence in computer-mediated communications.
We propose a method to learn emoji representations called EmoGraph2vec, which can also be utilized in other NLP tasks. The feture fusion hybrid-attention network we designed learns the mutual emotional semantics between text and emojis on texts of social media.


\section*{Acknowledgment}
The research is supported in part by the Strategic Priority Research Program of the Chinese Academy of Sciences, Grant No. XDC02060400.




%

\end{document}